\documentclass[runningheads]{llncs}
\usepackage{graphicx}
\begin{document}
\title{L3Cube-HingCorpus and HingBERT: A Code Mixed Hindi-English Dataset and BERT Language Models \thanks{Supported by L3Cube Pune.}}
\titlerunning{L3Cube-HingCorpus and HingBERT}
%
\author{Ravindra Nayak\inst{1,3} \and
Raviraj Joshi\inst{2,3}
}
\authorrunning{R. Nayak and R. Joshi}
%
\institute{Sri Jayachamarajendra College of Engineering, Mysuru, Karnataka, India \and
Indian Institute of Technology Madras, Chennai, Tamilnadu India
\and
L3Cube, Pune\\
\email{\{ravirajoshi,ravindranyk707\}@gmail.com}}
\maketitle              
\begin{abstract}
Code-switching occurs when more than one language is mixed in a given sentence or a conversation. This phenomenon is more prominent on social media platforms and its adoption is increasing over time. Therefore code-mixed NLP has been extensively studied in the literature. As pre-trained transformer-based architectures are gaining popularity, we observe that real code-mixing data are scarce to pre-train large language models. We present L3Cube-HingCorpus, the first large-scale real Hindi-English code mixed data in a Roman script. It consists of 52.93M sentences and 1.04B tokens, scraped from Twitter. We further present HingBERT,  HingMBERT, HingRoBERTa, and HingGPT. The BERT models have been pre-trained on codemixed HingCorpus using masked language modelling objectives.  
We show the effectiveness of these BERT models on the subsequent downstream tasks like code-mixed sentiment analysis, POS tagging, NER, and LID from the GLUECoS benchmark. The HingGPT is a GPT2 based generative transformer model capable of generating full tweets.
We also release L3Cube-HingLID Corpus, the largest code-mixed Hindi-English language identification(LID) dataset and HingBERT-LID, a production-quality LID model to facilitate capturing of more code-mixed data using the process outlined in this work.
The dataset and models are available at {\footnotesize https://github.com/l3cube-pune/code-mixed-nlp}.

\keywords{code mixed \and BERT \and code switch \and Hinglish \and English \and Hindi \and MBERT \and XLM-RoBERTa \and HingCorpus \and HingBERT \and GPT.}
\end{abstract}

\section{Introduction}

Popular languages like English have been penetrating non-English societies. The usage of English along with other local languages has drastically increased. As people are getting accustomed to it, there is also a need of understanding such code-mixed data. In this internet era, we see the usage of code-mixed data prevalently in social media and chat platforms \cite{kim2006reasons}. We observe that there is a mismatch between the scale at which this code-mixed language is used and the data that is available for further research. 

As Hindi is the third most spoken language in the world after English and Mandarin\footnote{\scriptsize\url{https://en.wikipedia.org/wiki/Hindi}}. The usage of Hinglish, a portmanteau of Hindi and English \cite{srivastava2021challenges,gupta2020semi} has become popular in the recent past in the Indian sub-continent. Since it is difficult to build a large scale code-mixed dataset, the literature has been more inclined toward building synthetic code-mixed datasets \cite{srivastava2021hinge}. However, at the same time real code-mixed data has been shown to produce better results than synthetically generated datasets \cite{santy2021bertologicomix}. We, therefore, aim to build a real Hinglish data corpora which can be used to enhance other code-mixed NLP tasks. In this work, we build L3Cube-HingCorpus a Hindi-English code-mixed corpus, containing 52.93M sentences and 1.04B tokens.

The unsupervised HingCorpus is further used to train BERT based language models. The BERT based architectures have gained traction recently, due to their various pre-training and fine-tuning techniques that have taken over initial deep learning techniques. The unsupervised pretraining has shown promising results on deep neural network architectures as they act like a regulariser to the model \cite{JMLR:v11:erhan10a}. So we pre-train the model on the masked language modelling task and then further try to evaluate various downstream tasks.

We introduce transformer-based BERT models \cite{devlin2019bert}, namely HingBERT\footnote{\scriptsize\url{https://huggingface.co/l3cube-pune/hing-bert}},
HingMBERT\footnote{\scriptsize\url{https://huggingface.co/l3cube-pune/hing-mbert}}\footnote{\scriptsize\url{https://huggingface.co/l3cube-pune/hing-mbert-mixed}}, 
and HingRoBERTa\footnote{\scriptsize\url{https://huggingface.co/l3cube-pune/hing-roberta}}\footnote{\scriptsize\url{https://huggingface.co/l3cube-pune/hing-robera-mixed}} 
all pre-trained on our Hinglish corpus. We release both roman and mixed script versions of these models trained on roman script text and roman + Devanagari text respectively. The models have been evaluated on various downstream tasks such as Language Identification(LID), Named Entity Recognition(NER), Part of Speech(POS) tagging and Sentiment analysis, which were part of the GLUECos benchmark dataset \cite{khanuja2020gluecos}. We also release other resources like HingGPT
\footnote{\scriptsize\url{https://huggingface.co/l3cube-pune/hing-gpt}}\footnote{\scriptsize\url{https://huggingface.co/l3cube-pune/hing-gpt-devanagari}}
, a GPT2 \cite{radford2019language} model trained on HingCorpus and HingFT, the fast text \cite{mikolov2018advances} based code-mixed Hindi-English word embeddings. 

To facilitate further creation of code-mixed Hi-En corpus we release HingBERT-LID
\footnote{\scriptsize\url{https://huggingface.co/l3cube-pune/hing-bert-lid}}
, a token level Hindi-English language identification model trained on a large in-house LID dataset. The model can be utilized to select code-mixed Hi-En sentences and expand the HingCorpus using the process outlined in the paper.  
A subset of the LID dataset is released as a benchmark code mixed Hindi-English language identification dataset L3Cube-HingLID. This is the largest LID dataset for the Hi-En pair. 

The data and models is publicly\footnote{\scriptsize\url{https://github.com/l3cube-pune/code-mixed-nlp}} 
released to enable further research in Hinglish NLP.

\section{Related Work}
In this section, we will try to mainly discuss previous attempts in the creation of code-mixed datasets. User-generated content is the main source of code-mixed data, and preprocessing is necessary for tasks like profanity hate speech \cite{qin2020cosdaml,bohra2018dataset,kamble2018hate,santosh2019hate,nayak2021contextual}, sentiment analysis, etc. Various attempts of scraping have been done before for the initial set of code-mixed data and later augmented synthetically using equivalence constraint theory \cite{pratapa-etal-2018-language}, semi-supervised learning \cite{gupta-etal-2020-semi} and rule-based language-pair approaches \cite{srivastava2021hinge}.

As BERT based architectures are gaining popularity, there have been studies around pre-training and fine-tuning them on various tasks. There have been variations around the BERT architecture like RoBERTa \cite{liu2019roberta} and ALBERT \cite{lan2020albert}, which have helped in various use cases like accuracy and latency related improvements. Models like multilingual-BERT, XLM-RoBERTa \cite{conneau2020unsupervised}, have focused mainly on multilingual and cross-lingual data representations. 

While evaluating code-mixed tasks, it is also shown that training on code-mixed sentences has given better results compared to training them on multiple monolingual corpora \cite{ansari2021language}. Bertlogicomix \cite{santy2021bertologicomix} have shown that real code-mixed data works much better when compared to synthetically generated after fine-tuning on various BERT based architectures. All the above models have been pre-trained on not more than 100k real code-mixed sentences. GLUECoS, the benchmark dataset was also evaluated on models pre-trained using  5M sentences which was a mix of both real and synthetic code-mixed sentences \cite{khanuja2020gluecos}.

\section{Curation of Dataset}

Our data consists of tweets that were scraped using the framework Twint \footnote{\scriptsize\url{https://github.com/twintproject/twint}}. An initial vocabulary of commonly spoken Hindi words was iteratively built to scrape the tweets containing these words. The initial vocabulary was constantly updated to include the newly found words from the scraped data. As we focus only on the code-switched Roman script, the scraped data was then preprocessed to remove non-English characters. User mentions in the tweet were also removed to avoid privacy concerns.

The pre-processed data is passed through a word-level language classifier model to detect the language of each word. If both Hindi and English words are present in the sentence it is treated as a code-mixed sentence. The language classifier is initially a shallow subword-based LSTM as described in \cite{joshi2022evaluating}. The shallow model is shown to work well for Hindi-English language identification on limited data. The model is trained iteratively using a semi-supervised learning approach. A small labelled dataset with 5k sentences was created initially and further multiple versions of the models are trained using the pseudo-labels generated from the previous version. We manually verified less confident pseudo labels and corrected labels were fed for the next iteration of training. In the end, we create a dataset of around 44455 sentences using this process. Finally, the expanded dataset is used to fine-tune the base BERT model as it worked better than the LSTM counterpart. This ensured that we have a strong word language classifier in place while creating the target dataset. It was ensured that the LID model was highly accurate as the quality of the Hinglish corpus depended heavily on the LID accuracy. The details of the LID accuracy are discussed in the results section. We set a threshold to check whether a sufficient number of Hindi and English words are present in the sentence to consider it as code-mixed. A sentence is considered code-mixed if it has at least 2 Hindi and 2 English words. We have retained the case, punctuation and smileys in the sentences and the data were shuffled in the end for training.

The final dataset consists of nearly 52.93M sentences (1.04B tokens), out of which 47.79M (944M tokens) sentences were used for training and 5.13M sentences (99M tokens) for validation. The Devanagari version of HingCorpus is created using an in-house transliteration model. The Devanagari dataset contains an equal number of sentences and an approximately similar number of tokens. A code-mixing metric viz Mixed CMI index \cite{GAMBCK16.669} of 31.21 was obtained from final data, where 0 corresponds to monolingual data with no code-mixing and 100 is the highest degree of code-switching.

\section{Model Architecture}
Our architecture includes various BERT model variations, that are trained on unsupervised learning tasks like masked language model (MLM) and next sentence prediction (NSP). Deep bi-directional transformers are the basic building block of these models. Their use has been prevalent due to their understanding of the long term dependencies of text. Moreover, they are capable of making use of contemporary hardware to train the models parallelly. We explore three variations of BERT-based models viz. BERT-base, m-BERT and XLM-RoBERTa. 

\begin{itemize}
\item \textbf{BERT :}
Also known as BERT-base \cite{devlin2019bert}, it is a model that contains 12 transformer blocks, 12 self-attention heads, hidden size of 768. The input for BERT contains a maximum embedding of 512 words and it outputs a sequential representation. Special tokens like [CLS] and [SEP] are used to specify the start of a sentence and separation of sentences respectively. For a classification task, final encoder representations are considered and a softmax is applied to classify the representation.

\item \textbf{Multilingual-BERT (m-BERT) :}
This model’s architecture is based on BERT-base. It has been trained in 102 languages with a word-piece vocabulary of size 110k \cite{devlin2019bert}. It has shown promising results for zero-shot transfer learning on various downstream tasks and also helped in code-switched data tasks \cite{pires2019multilingual}.

\item \textbf{XLM-RoBERTa :}
It is a transformer-based multi-lingual language model which has been trained on 100 languages \cite{conneau2020unsupervised}. It has shown great results in cross-lingual tasks and has outperformed m-BERT in various multi-lingual downstream tasks.
\end{itemize}

\section{HingBERT Evaluation}
\subsection{Training}
In this work, we consider three variations of BERT architectures i.e. BERT, m-BERT and XLM-RoBERTa for training. These models are further pre-trained on L3Cube-HingCorpus using MLM objective with a masking probability of 15\%. The models were trained for 2 epochs with a learning rate of 1e-5 and a batch size of 64. We observed that 2 epochs were sufficient for the models to converge as we loaded the pre-trained weights of the respective models. Moreover, there was no significant decrease in the loss after 2 epochs. The respective models after Hinglish training are referred to as HingBERT, HingMBERT and HingRoBERTa and their validation perplexity on this task is shown in Table \ref{stats}. These were further fine-tuned on the respective downstream tasks by considering the [CLS] or token embeddings and feeding it to feed-forward layers. We train two versions of models in Roman script and mixed script. The mixed script model is trained on both roman and Devanagari text. The mixed script model can be used for both roman or Devanagari code-mixed text. The mixed script HingBERT models are evaluated on the Devanagari version of the GLUECoS dataset.

\begin{table}
\centering
\caption{This table shows the evaluation of pre-training the model on the MLM task. Perplexity is a measure to validate how well the language model can predict the next word, in the case of BERT it would be the prediction of masked words.}
\label{stats}
\begin{tabular}{c c}
\hline
    \textbf{Model} & \textbf{Validation Perplexity} \\
\hline
    HingBERT & 5.72 \\
    HingMBERT & 5.20 \\
    HingRoBERTa & 7.82 \\
    HingMBERT-mixed & 5.22 \\
    HingRoBERTa-mixed & 9.39 \\
\hline
\end{tabular}
\end{table}

\subsection{Downstream Tasks}
For the evaluation of our models, we use the EN-HI pair from GLUECoS, a code-switching benchmark dataset, for the below mentioned NLP tasks. The models were fine-tuned by adding a dense layer on top of the BERT encoder. These were fine-tuned for 5 epochs using early stopping w.r.t validation F1 score. A batch size of 64 and a learning rate of 3e-5 were used.

\begin{enumerate}
\item \textbf{Language Identification (LID)}: This task is to mainly identify the language for each word in the given sentence, with labels EN (English), HI (Hindi) and OTHER. This task contains 2631 training data points along with 500 dev and 406 test data, and the SOTA was achieved by the GLUECoS-mBERT model \cite{khanuja2020gluecos}.

\item \textbf{Part of Speech (POS) tagging}: There are 2 datasets under this subtask which are named POS-UD and POS-FG. POS-UD has 16 labels to predict with  1384 data points for training and 215 \& 215 for dev \& test respectively. Similarly, POS-FG has 2104, 263 \&  264 data points for training, dev and testing with nearly 35 unique labels. The highest score is mentioned state-of-the-art (SOTA) models for these tasks given by papers \cite{bhat-etal-2018-universal} for POS-UD and \cite{Sharma2015POSTF} for POS-FG.

\item \textbf{NER (Named Entity Recognition)}: This is a token level classification task for words consisting of 7 labels. There are 2467 training data sentences and 308 \& 309 sentences for validation \& testing. The SOTA was achieved by GLUECoS-mBERT model \cite{khanuja2020gluecos}.

\item \textbf{Sentiment analysis}: This is a multi-class classification task of predicting the sentiment of the sentence as positive, negative or neutral. This dataset contains 10080, 1260 and 1261 sentences for train, dev and test sets respectively. GLUECoS-mBERT model was able to achieve SOTA on this task.
\end{enumerate}

\section{L3Cube-HingLID Corpus}
The LID dataset used to train the LID model is termed L3Cube-HingLID and is released publicly as the benchmark dataset. The L3Cube-HingLID consists of 31756, 6420, and 6279 train, test, and validation samples respectively with an average of nearly 30 tokens per sentence across all the datasets. All the models considered in this work are also evaluated on this LID dataset. Note that the HingBERT-LID model released as a part of this work was trained on a bigger corpus and provides the best numbers on the L3Cube-HingLID test set as compared to the models trained only on its train set. It was ensured that the test and validation set were separate and not leaked during the training of HingBERT-LID. The original LID train set was further expanded using the first generation of the BERT model trained on this train set to label an equal amount of unlabelled datasets. Both the supervised data and unsupervised data were used to train the final model. This strong LID model with 98\% of accuracy on the unseen test set was used for selecting sentences for HingCorpus.  

\begin{table}
\centering
\caption{This table shows the token level details of HingLID dataset}
\label{stats}
\begin{tabular}{c c c}
\hline
    \textbf{Data} & \textbf{EN} & \textbf{HI}\\
\hline
    Train & 274255 & 693977 \\
    Test & 56723 & 136824 \\
    Validation & 56143 & 137575 \\
\hline
\end{tabular}
\end{table}

\section{Other Resources}
\subsection{HingGPT}
HingGPT is a standard GPT2 causal transformer model trained on HingCorpus using the language modelling task. The model has 12 standard transformer layers and is trained using the Causal Language Model (CLM) objective. With a learning rate of 5e-5, the model is trained for 2 epochs. We train both roman and Devanagari versions of the model and are capable of generating full tweets. The mixed script version is not relevant for GPT and hence is not considered. The model can be further used to either generate or evaluate the quality of synthetic code-mixed corpus. Some sample tweets generated using roman HingGPT are shown in Table \ref{hing_gpt_sample}.

\subsection{HingFT}
We train fast text style distributed word representations using the HingCorpus and term it as HingFT. A skip-gram model is used to train 300 dimension word embeddings using the standard training parameters. The model is trained for 10 epochs using a learning rate of 0.05. The fast text uses subwords to create word embeddings and is more suitable for code-mixed text.  
\begin{table*}
\centering
\caption{This table shows some of the sentences generated by our HingGPT model. The words that are in bold are the initial text provided to the model to generate the sentences}
\label{hing_gpt_sample}

\begin{tabular}{c}
\hline
    \textbf{Sentences generated by HingGPT} \\
\hline
    \textbf{My name is Julien and I like to} make food for you every hour and \\ see the whole world without you , I can ' t even keep you happy , \\ I know you will be missed You are the best but how do you forget to add , \\the world is the best , it has the ultimate universe .\\
    \hline
    \textbf{mujhe iss duniya} se  jana , na ki zindagi se . mujhe bas apna rehna ... \\ teri ek muskan se bhi mangi hui har kami ko hai . \\ dil mein hai it ' s so weird to even see people who ask \\ for their rights are just asking to follow .\\
    \hline
    \textbf{The goal of life} is not to lose trust of your own self . \\ And it ' s more important than your own self .\\
    \hline
    \textbf{The goal of life is} not merely a mere lawyers document , \\it is a vehicle of life , and its spirit is always the spirit of age . - Dr . Khan\\
    \hline
    \textbf{Corona has} become worse . So , for now , for the benefit of the family , \\ we have to pay enough to get our daughters vaccinated . \\ If we had a booster , it ' s better our kids should ' ve.\\
    
\hline
\end{tabular}
\end{table*}

\begin{table*}
\centering
\caption{This table represents the F1 scores of test sets after fine-tuning on various downstream tasks of the GLUECoS dataset in Roman script}
\label{results_roman}
\begin{tabular}{c c c c c c c}
\hline
    \textbf{Model} & \textbf{LID} & \textbf{POS-UD} & \textbf{POS-FG} & \textbf{NER} & \textbf{Sentiment} & \textbf{HingLID}  \\

\hline
    BERT & 78.69 & 83.70 & 70.75 & 79.27 & 59.16 & 96.04\\
    m-BERT & 82.56 & 83.68 & 69.58 & 76.64 & 58.42 & 95.59\\
    XLMRoBERTa & 85.93 & 87.24 & 70.95 & 77.01 & 61.57 & 95.42\\

\hline
    HingBERT & 84.44 & 88.42 & 71.04 & \textbf{81.80} & 63.72 & 96.21\\
    HingMBERT & 84.90 & 89.47 & 71.55 & 80.09 & 63.51 & 96.27\\
    HingRoBERTa & \textbf{86.69} & \textbf{90.17} & \textbf{71.69} & 81.13 & 66.43 & 96.15\\
\hline
    HingMBERT-mixed & 83.26 & 90.06 & 70.34 & 81.12 & 63.51 & 96.29\\
    HingRoBERTa-mixed & 86.13 & 89.87 & 70.73 & 80.68 & \textbf{66.73} & 95.96\\
    HingBERT-LID & - & - & - & - & - & \textbf{98.77}\\
\hline
\end{tabular}
\end{table*}

\begin{table*}
\centering
\caption{This table represents the F1 scores of test sets after fine-tuning on various downstream tasks of the GLUECoS dataset in mixed script including Roman and Devanagari script. }
\label{results_devanagari}
\begin{tabular}{c c c c c c}
\hline
    \textbf{Model} & \textbf{LID} & \textbf{POS-UD} & \textbf{POS-FG} & \textbf{NER} & \textbf{Sentiment}\\
\hline
    SOTA & 96.6 & 90.53 & 80.68 & 78.21 & 59.35\\
\hline
    BERT & 95.30 & 81.49 & 68.55 & 73.92 & 60.14\\
    m-BERT & 95.03 & 86.87 & 69.81 & 74.79 & 60.45\\
    XLMRoBERTa & 95.37 & 89.62 & 70.53 & 75.53 & 63.93\\
    
\hline
    GLUECoS-mBERT & 96.6 & 88.06 & 63.31 & 78.21 & 59.35\\
    BERToLogicoMix & 95.8 & 88.09 & 60.46 & 76.86 & 58.25\\

\hline
    HingBERT & 95.54 & 82.26 & 67.69 & 77.60 & 59.59\\
    HingMBERT & 95.68 & 86.71 & 70.15 & 78.78 & 60.72\\
    HingRoBERTa & \textbf{96.30} & 89.97 & 69.90 & 80.28 & 64.43\\
\hline
    HingMBERT-mixed & 95.65 & 89.31 & 70.52 & 79.66 & 62.93\\
    HingRoBERTa-mixed & 94.96 & \textbf{90.81} & \textbf{70.61} & \textbf{81.72} & \textbf{66.07}\\

\hline
\end{tabular}
\end{table*}

\subsection{Results and Discussions}
All the HingBERT models are pre-trained with similar hyper-parameters and fine-tuned on different tasks. The models are evaluated on 3 token classification tasks POS, NER, LID and one sentence classification task of sentiment identification. These tasks are part of the GLUECoS benchmark.  We use the F1 score as the metric for the evaluation of these models. The dataset for these tasks is present in roman and mixed script form. The mixed script is mostly in the Devnagari script along with some roman tokens. The results for all the tasks in Roman script are described in Table \ref{results_roman}. Table \ref{results_devanagari} describes the results for tasks in mixed Devanagari + Roman script. Along with models introduced in this work we also evaluate baseline models like base BERT, m-BERT, and XLMRoBERTa. Table \ref{results_devanagari} also shows various SOTA F1 scores on all these tasks. The mixed script form of the dataset has been mainly evaluated in the literature so SOTA numbers are only added for the mixed script form.  We observe that our models outperform SOTA numbers on NER and Sentiment tasks. They perform competitively on the LID and POS-UD tasks. They perform poorly only on the POS-FG mixed-script task where all the BERT models fail to compete with SOTA. However, our models consistently outperform the baseline BERT models on all the tasks. Both roman and mixed-script models perform better than their respective baselines on either of the script. We see that the roman models perform slightly better than the mixed script ones on the roman script tasks. Similarly, mixed-script models perform better than roman models on mixed-script tasks. The observations are consistent with the general assumption that the addition of Devanagari data will help the mixed-script tasks containing Devanagari words. Among the models introduced in this work, the RoBERTa based models mostly perform the best. It outperforms all the BERT based models and also achieves SOTA on three tasks except for the POS-FG and LID tasks. Overall we show that pre-training on real code-mixed corpus provides significant performance improvements. 

\section{Conclusion}
In this paper, we expand the code mixed Hindi-English corpora, using various data mining and curation techniques. We present L3Cube-HingCorpus, the first major unsupervised Hindi-English code-mixed dataset. We have used these corpora in pre-training our BERT based models namely HingBERT, HingMBERT, HingRoBERTa. These models were later evaluated on various downstream NLP tasks. We observe that pretraining the models on real code mixed data have helped them outperform BERT models pre-trained on non-code-mixed corpus and synthetic code-mixed corpus and achieve SOTA on the majority of these tasks. We also release other resources like HingGPT, a GPT2 model and HingFT, a Hinglish fast-text model both trained on HingCorpus. We leave the evaluation of these models on downstream tasks to future work. Finally, we curate a new Hindi-English LID Corpus HingLID containing around 44k sentences and also release HingBERT-LID to further help augmentation of HingCorpus.

\section*{Acknowledgements} This work was done under the L3Cube Pune mentorship program. We would like to express our gratitude towards our mentors at L3Cube for their continuous support and encouragement.

\bibliographystyle{splncs04}
\bibliography{main}
%




\end{document}